 \documentclass[letterpaper, 10 pt, journal, twoside]{IEEEtran}

\IEEEoverridecommandlockouts

\usepackage{amsmath, amssymb}
\usepackage{stfloats}
\usepackage{physics}

\usepackage{graphics} 
\usepackage{epsfig} 
\usepackage{times} 
\usepackage{amsmath} 
\usepackage{amssymb}  
\usepackage[table]{xcolor}
\usepackage[ruled, linesnumbered, noend]{algorithm2e}
\SetKwRepeat{Do}{do}{while}
\usepackage{adjustbox}
\usepackage{booktabs}
\usepackage{siunitx}

\usepackage{tablefootnote}
\usepackage{lipsum} 
\usepackage{soul}
\usepackage{cite}
\usepackage{comment}
\usepackage{multirow}
\usepackage{graphicx}
\usepackage{wrapfig}
\usepackage{subfigure}
\usepackage{upgreek}
\usepackage{diagbox}

\usepackage{romannum}
\usepackage{pifont}

\makeatletter
\newcommand{\removelatexerror}{\let\@latex@error\@gobble}
\makeatother

\title{Motion Planning for Minimally Actuated\\ Serial Robots}

\author{Avi Cohen, Avishai Sintov and David Zarrouk
\thanks{A. Cohen and D. Zarrouk are with the Department of Mechanical Engineering, Ben-Gurion University of the Negev, Israel.}
\thanks{A. Sintov is with the School of Mechanical Engineering, Tel-Aviv University, Israel. e-mail: \small sintov1@tauex.tau.ac.il.}
}

\newcommand{\ve}[1]{\mathbf{#1}} 
\newcommand{\tve}[1]{\tilde{\mathbf{#1}}} 
\newcommand{\mr}[0]{MASR-RRT* } 
\newcommand{\mrr}[0]{MASR-RRT*} 

\begin{document}

\setlength{\belowdisplayskip}{4pt}
\setlength{\belowdisplayshortskip}{3pt}
\setlength{\abovedisplayskip}{4pt} 
\setlength{\abovedisplayshortskip}{3pt}
\setlength{\parskip}{0pt}


\maketitle
\thispagestyle{empty}
\pagestyle{empty}

\begin{abstract}
Modern manipulators are acclaimed for their precision but often struggle to operate in confined spaces. This limitation has driven the development of hyper-redundant and continuum robots. While these present unique advantages, they face challenges in, for instance, weight, mechanical complexity, modeling and costs. 
The Minimally Actuated Serial Robot (MASR) has been proposed as a light-weight, low-cost and simpler alternative where passive joints are actuated with a Mobile Actuator (MA) moving along the arm. Yet, Inverse Kinematics (IK) and a general motion planning algorithm for the MASR have not be addressed. In this letter, we propose the \mr motion planning algorithm specifically developed for the unique kinematics of MASR. The main component of the algorithm is a data-based model for solving the IK problem while considering minimal traverse of the MA. The model is trained solely using the forward kinematics of the MASR and does not require real data. With the model as a local-connection mechanism, \mr minimizes a cost function expressing the action time. In a comprehensive analysis, we show that \mr is superior in performance to the straight-forward implementation of the standard RRT*. Experiments on a real robot in different environments with obstacles validate the proposed algorithm.

\end{abstract}


\section{Introduction}
\label{sec:introduction}

Robots are often required to work in cluttered or confined spaces in applications such as spacecraft maintenance \cite{Peng2021a}, oil and gas pipelines \cite{Peng2021}, search and rescue \cite{Wolf2005} and minimally invasive procedures \cite{BurgnerKahrs2015}. However, conventional serial robots that excel in accuracy and velocity, struggle to operate in such environments \cite{Mu2022}. In order to cope with the limitations, hyper-redundant manipulators have been developed \cite{hyper_redundant}. These manipulators consist of a high number of Degrees-Of-Freedom (DOF) in order to exert high flexibility and maneuverability in confined spaces. Nevertheless, they can be highly expensive due to the large number of actuators, are not power efficient, and have a low payload. Continuum robots, on the other hand, are lighter and require fewer actuators due to their underactuated compliant mechanism \cite{Qin2022,continuum_robot2}. However, these mechanisms often have a complex structure with highly nonlinear behavior, yielding difficulties in kinematic modeling and motion planning along with low payload capacity \cite{russo2023continuum}.


\begin{figure}[h]
\centering
\includegraphics[width=\linewidth]{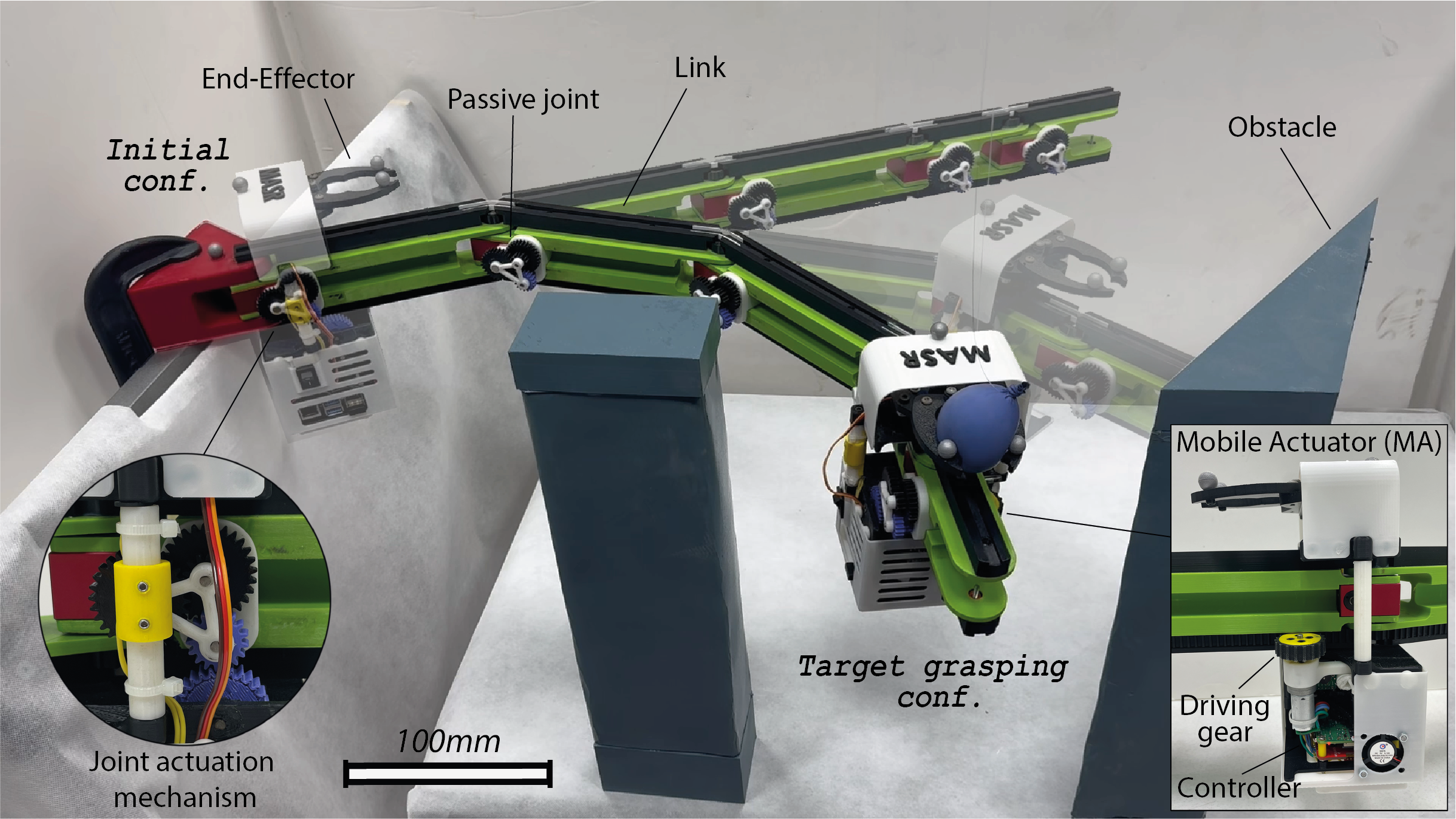}
\vspace{-0.7cm}
\caption{The Minimally actuated serial robot (MASR) with its serial design and a Mobile Actuator (MA), moving from an initial straight configuration to grasping a balloon near obstacles (see video).}
\label{fig:masr}
\vspace{-0.7cm}
\end{figure}
In previous work, a novel class of serial robots, termed \textit{Minimally Actuated Serial Robot} (MASR), was proposed \cite{Mann2018,2dmasr}. The MASR, seen in Fig. \ref{fig:masr}, is a multi-joint, lightweight and modular robot arm with a minimal number of actuators. It comprises a serial set of rigid links and passive joints where a single active Mobile Actuator (MA) travels along the arm to actuate selected joints. Unlike conventional manipulators, the End-Effector (EE) of the MASR is mounted on the MA. Thus, the redundancy of the arm is augmented by the ability to move a gripper along it. Due to the MASR's unique design, its control is simple, even with an increase in the number of passive joints. 

Although the design of MASR is advantageous, two algorithmic issues remain to be addressed. First, an optimal solution to the specific Inverse kinematics (IK) problem of MASR is required. In particular, an IK solution for a desired configuration of the MASR requires consideration of an optimal MA motion to sequentially actuate joints along with the corresponding final pose. Approaches for IK are predominantly divided into analytical and numerical methods. The former offers a closed-form solution while being usually infeasible for complex and redundant robots \cite{lynch2017}. On the other hand, numerical methods, commonly based on least squares and gradient descent, cater to a broader spectrum of kinematic chains \cite{levenberg,jacobian}. However, the methods suffer from inconsistent convergence rates and high dependency on the initial guess. Over the years, data-based learning approaches have been explored to solve complex IK problems \cite{ik_nn_2001,ik_nn_2004}. In particular, Neural Networks (NN) have been shown to handle complex kinematics and provide real-time inference \cite{ik_nn_2017,ik_nn_2020, ik_nn_2022}. Nevertheless, all data-based IK approaches are designated for conventional manipulators and cannot provide a solution to MASR's unique kinematics. 

The second algorithmic issue to be explored is the motion planning problem. Due to its unique kinematics and passive joints, the MA must reach a joint to actuate it, making MASR slower than conventional manipulators. Hence, the MASR requires an optimal path in terms of minimum actuation of joints and MA traverse time. Asymptotically optimal sampling-based motion planning algorithms, such as the optimal Rapidly Exploring Random Trees (RRT*) \cite{Karaman2011} can address this. However, these standard algorithms can easily offer motion planning in the joint space of the arm while not being able to consider optimal sequences of MA motions between joints. Early work demonstrated a simple and non-optimal motion planning for the MASR by attempting to track the motion of a fully actuated arm \cite{Mann2018}. However, optimal motion of the MA was not considered. In a more recent work, a planning algorithm was proposed for optimal kinematic designs of a MASR for a specific task  \cite{yanai2022}. To the best of the author's knowledge, a general, comprehensive and optimal motion planning algorithm for the MASR has yet to be proposed.

In this letter, we address the above challenges and propose a novel motion planning algorithm, termed \mrr, specifically developed for MASR. \mr consists of an IK solver based on a NN (IK-NN). IK-NN does not provide some configuration solution to a desired EE pose but an optimal one in terms of minimal traverse time from the current configuration for both the joints and MA. Furthermore, training the IK-NN does not require tedious collection of data from a real or simulated robot, but simply through the computation of the forward kinematics. A designated regularization term specifically tailored for MASR is minimized during the training of IK-NN to reduce joint rotation and MA movements along the arm. In such approach, MASR exploits the advantage of the MA, where the MA can be positioned anywhere along the arm, yielding less joint actuation. Hence, IK-NN is a primary component of the \mr algorithm. 

\mr is an adaptation of the RRT* for MASR where standardization of the MA motion is defined. In addition, the IK-NN is stochastically used in order to form local connections within the search tree with bias toward the goal. \mr optimizes a cost function reflecting the action time of the robot along the path. With the modified features, \mr is able to better converge to the optimal solution path compared to the standard RRT*. We present a thorough evaluation of planning performance along with real robot demonstrations. 
\vspace{-0.3cm}

\section{Minimally Actuated Serial Robot}
\label{sec:masr}

This section overviews the operation and kinematics of MASR, and defines the motion planning problem. The MASR is composed of a completely passive serial arm without any electrical components and a single active Mobile Actuator (MA). MASR is primarily manufactured from 3D-printed materials and is illustrated in Fig. \ref{fig:masr_and_scheme}a. The MA runs along a designated track on the serial arm to reach designated joints and actuate them. In such a way, the MA can change the configuration of the robot. For detailed information on the design and control, readers may refer to previous work \cite{2dmasr}.


\subsection{Design and Actuation}
\label{subsec:design_actuation}

The MASR arm is an open kinematic chain consisting of $n$ links with a total length $l=\sum_{j=1}^{n} l_{j}$ where $l_j$ is the length of the $j$-th link. The links are connected through $n$ passive revolute joints, each constructed from a worm gear transmission to maintain angle lock when not actuated. The relative angle of joint $j$ between two adjacent links $l_{j-1}$ and $l_j$ is denoted by $\theta_{j}$ and can vary within the range $\left[ -\theta_{j}^b, \theta_{j}^b \right]$. The rotational speed of the joint is denoted by $\dot{\theta}_j$. The one-dimensional position and velocity of the MA along the arm are defined by $d \in[0,l]$ and $\dot{d}$, respectively. The MA can rotate the $j$-th joint only if $d=r_j\in\mathbb{R}$ where $r_j=\sum_{i=1}^{j-1} l_{i}$.

\begin{figure}[h]
\vspace{-0.3cm}
\centering
\includegraphics[width=88mm]{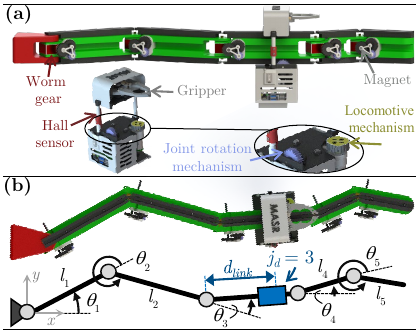}
\vspace{-0.6cm}
\caption{(a) MASR components, serial arm and mobile actuator (MA) with its mechanisms. (b) A schematic diagram showing a typical $5$R MASR configuration $\mathbf{q}$, with an example of how $j_{d}$ and $d_{link}$ are derived from $d$ (i.e., $d=r_3+d_{link}=l_1+l_2+d_{link}$).}
\label{fig:masr_and_scheme}
\vspace{-0.7cm}
\end{figure}

The MA has two controlled mechanisms. The \textit{locomotive mechanism} advances the MA along the links by rotating a wheel with a single actuator. The position of the MA is controlled using an encoder on the wheel and a hall sensor that senses the arrival of magnets attached to the joints. By accumulating the passages over the magnets and with the encoder, the position of the MA is calculated by $d = r_{j_d} + d_{link}$ where $j_{d}$ is the adjacent joint index and $d_{link}$ is the relative position along corresponding link $j_d$ (see Fig. \ref{fig:masr_and_scheme}b). Velocity $\dot{d}$ is regulated by a PID controller, which uses the click rate of the motor encoder. The second mechanism, termed \textit{joint rotation mechanism}, actuates a specific joint $j$ at position $d=r_j$. The MA stops on joint $j$ when reaching its magnet, followed by rotations of a spur gear actuated by a single actuator. The gear is controlled by an encoder and engages the spur gear of the joint. As a result, the worm gear rotates and changes the joint angle $\theta_{j}$ with velocity $\dot{\theta}$. Angle and velocity are regulated through a PID controller that uses the encoder information. On top of these, a gripper is mounted on the MA in order to complete tasks and is actuated by a servo actuator.

\subsection{Forward Kinematics}
\label{subsec:kinematics}

The configuration space of a typical $n$-DOF serial manipulator is a subset of $\mathbb{R}^n$ and is defined by the set of all possible joint values. For the MASR, however, an additional DOF is included being the one-dimensional position $d$ of the MA along the arm. Therefore, the configuration space $\mathcal{C}\subset\mathbb{R}^{n+1}$ of MASR is the set of all joint positions and the position of the MA. Consequently, a configuration $\ve{q}\in\mathcal{C}$ is defined as 
\begin{equation}
    \label{eq:q}
    \ve{q} = (\boldsymbol{\uptheta}^T, d)^T
\end{equation}
where $\boldsymbol{\uptheta}=\{\theta_1,\ldots\theta_n\}$.

The Forward Kinematics (FK) function $f:\mathcal{C}\to SE(2)$ maps a configuration $\ve{q}$ to the pose of the gripper $\ve{x} \in SE(2)$. The function is given by
\begin{equation}
    \mathbf{x} = f(\mathbf{q}) = 
    \begin{bmatrix}   
    \multicolumn{2}{c}{\multirow{2}{*}{$R(\hat{\theta}_{j_d})$}}    & x \\
    & & y \\
    0 & 0 & 1 \\
    \end{bmatrix}
\end{equation}
where $\hat{\theta}_k=\sum_{j=1}^{k}\theta_{j}$, $R\in SO(2)$ is a rotation matrix about the axis perpendicular to the surface of motion, and
\begin{equation}
x = \sum\limits_{j=1}^{j_{d}-1}\left(l_{j}\cos\hat{\theta}_j\right)+
    d_{link}\cos\hat{\theta}_{j_d},
\end{equation}
\begin{equation}
y = \sum\limits_{j=1}^{j_{d}-1}\left(l_{j}\sin\hat{\theta}_j\right)+
    d_{link}\sin\hat{\theta}_{j_d}.
\end{equation}
For convenience, we denote $\ve{p}=(x,y)$ and $\phi=\hat{\theta}_{j_d}$ to be the position and orientation, respectively, of the gripper with respect to some global coordinate frame.


\subsection{Problem Definition}
\label{sec:problem_def}
Let $\mathcal{C}_{f}\subset\mathcal{C}$ be a feasible subset of the configuration space
\begin{equation}
    \mathcal{C}_{f} = \{\ve{q}\in\mathcal{C} : \theta_j \in [-\theta_{j}^b, \theta_{j}^b] \ \forall j \leq n \ \wedge \ d \in  [0, l]\}.
\end{equation}
Hence, the MASR workspace set $\mathcal{W}$ is obtained according to
\begin{equation}
    \mathcal{W} = \{\mathbf{x} \in SE(2) : \mathbf{x} = f(\mathbf{q}) \ \forall\mathbf{q} \in \mathcal{C}_{f}\}.
\end{equation}
Subset $\mathcal{C}_{b}$ is the restricted region due to obstacles in the environment. This includes the inflation of obstacles as seen in Fig. \ref{fig:alg_illustration}a. Hence, let the free configuration space of the MASR be $\mathcal{C}_{free}=\mathcal{C}_f \setminus \mathcal{C}_b$, which is the set of all configurations that satisfy the MASR constraints and are collision-free. 

The robot is situated in an initial configuration $\ve{q}_{init}\in\mathcal{C}_{free}$ and must reach a goal configuration $\ve{q}_{goal}\in\mathcal{C}_{free}$. The goal pose is derived from the goal configuration as $\mathbf{x}_{goal}=f(\mathbf{q}_{goal})$, or vice versa. 
A motion planning algorithm should find a path $\sigma:[0,1]\to\mathcal{C}_{free}$ such that $\sigma(0)=\ve{q}_{init}$ and $\sigma(1)=\ve{q}_{goal}$. A transition at step $s_k\in[0,1]$ from $\sigma(s_k)=\ve{q}_k$ to $\sigma(s_{k+1})=\ve{q}_{k+1}$ ($s_k<s_{k+1}$) is obtained by exerting action $\ve{a}_k=\left(\Delta \theta_{k,1},\ldots,\Delta \theta_{k,n}, \Delta d_k \right)^T=\ve{q}_{k+1}-\ve{q}_{k}\in\mathbb{R}^{n+1}$. 

An action $\ve{a}$ states which joints should be actuated in order to move from a current configuration $\ve{q}_c$ to a desired one $\ve{q}_d$. However, the action does not provide the order in which to move the MA between the joints. Therefore, we define a motion convention as follows. 
Let $d_c$ and $\Delta d\in\ve{a}$ be the current position and required step of the MA, respectively. Hence, the final MA position after the action should be $d_c+\Delta d$. 
If $\Delta d>0$, the MA will first move to the joints that satisfy $r_j<d_c$ (i.e., away from the final MA position) and then to those of $r_j>d_c$ (i.e., towards the final MA position). Otherwise, the order will be the opposite. In this way, the MA will move in an orderly fashion and not reach distant joints that do not require actuation.   
The total length to be moved by the MA following action $\ve{a}$ is denoted by $D_\ve{a}$. The computation of $D_\ve{a}$ is rather simple and demonstrated in Fig. \ref{fig:alg_illustration}b.

With the above definitions, the motion planning problem is to find an optimal path $\sigma^*(s)$ with $K$ steps $s=\{0,s_1,\ldots,s_{K-1},1\}$ that minimizes the overall action time $\tau$ calculated by
\begin{equation}
    \tau = \sum_{k=0}^{K-1} c(\ve{a}_k) 
\label{eq:time_action}
\end{equation}
where
\begin{equation}
    c(\ve{a}_k) = \frac{1}{\Dot{d}} D_{\ve{a}_k} +  \frac{1}{\Dot{\theta}}  \sum_{i=1}^n|\Delta \theta_{k,i}|.
\label{eq:one_action}
\end{equation}
is the cost of applying some action $\ve{a}$ to move between two configurations.

\subsection{Goal Reach Metric}
\label{subsec:metrics}

Let $\Delta {p} = \|\tve{p}-\ve{p}_{goal}\|_2$ and $\Delta \phi = |\Tilde{\phi}-\phi_{goal}|$ be the Euclidean distance and angular error of the estimated MA pose $\tve{x}$ with respect to the goal pose $\ve{x}_{goal}$. We define the subset $\mathcal{W}_{goal} \subset  \mathcal{W}$ as the region of poses considered to reach the goal according to
\begin{equation}
    \mathcal{W}_{goal} = \{\tve{x} \in\mathcal{W} : \Delta {p} \leq  e_p \wedge \Delta \phi \leq  e_{\phi} \},
\label{eq:goal_set}
\end{equation}
where $e_p\in\mathbb{R}$ and $e_{\phi}\in\mathbb{R}$ are pre-defined acceptable goal distance and angle errors, respectively.

\section{Inverse Kinematics Problem}
\label{sec:ik}
This section overviews a proposed method for solving the unique inverse kinematics (IK) problem $\mathbf{q}=f^{-1}(\mathbf{x})$ of MASR while minimizing the overall action time.




\subsection{Architecture}
\label{sec:optimal_tran_method}

To overcome the limitations of existing IK methods for the complex problem of MASR, we propose to optimize an IK solver using a Neural Network (NN). We aim to find an IK-NN model $g_\psi(\ve{x})\approx f^{-1}(\ve{x})$ where $\Psi$ consists of trainable model weights. An unsupervised learning approach is taken where the IK-NN model takes inspiration from an Autoencoder \cite{encoderdecoder}. An encoder-decoder architecture encodes the input data into a lower-dimensional representation and then decodes it back into the original or a slightly modified output format. Here, the encoder and decoder are IK $g_\psi$ and FK $f$, respectively. Once trained, the decoder is discarded and the trained encoder is an IK solver for the MASR. This approach eliminates the need for labeled data in a supervised learning strategy, while optimizing the motion to the IK solution.

\begin{figure}[h]
\includegraphics[width=88mm]{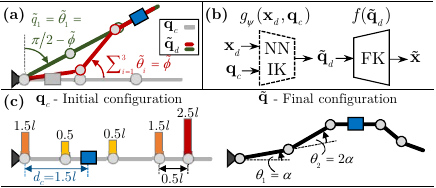}
\caption{(a) An example of redundancy in 3R MASR due to the additional MA DoF $d$ whereas the conventional 2D serial arm with $d=l$ has one solution on the last link tip. Here, the green configuration has a smaller overall action time than the red configuration since only the first joint must be actuated from the grey configuration. (b) The model architecture for training the IK-NN including a FK module. (c) Example of weights assigned to the joints of a 5R MASR by initial configuration $\ve{q}_c = (\mathbf{0}^T, 1.5l)^T$ and estimated configuration $\tve{q}=(\alpha, 2\alpha, -3\alpha, -3\alpha, 2\alpha, 2.5l)^T$ for $\mathcal{L}_{reg}$ calculation.}
\label{fig:ik_model_and_reg}
\vspace{-0.5cm}
\end{figure}

Model $\ve{q}=g_\psi(\ve{x})$ is trained by minimizing a loss function defined as
\begin{equation}
    \mathcal{L}=[\mathcal{V}_b]+\lambda\mathcal{L}_{reg},
    \label{eq:loss_mse_reg}
\end{equation}
where $\mathcal{V}_b=\log(\ve{x}_d^{-1}\tve{x})$ is a twist vector and $[V]$ is the skew-symmetric representation of a twist $V$ \cite{lynch2017}.
The first term in \eqref{eq:loss_mse_reg} states that the distance from the desired pose $\ve{x}_d$ and the predicted pose $\tve{x}=f(g_\psi(\ve{x}_d))$ should be minimized along the twist $\mathcal{V}_b\in\mathbb{R}^3$. The second term $\mathcal{L}_{reg}$ is a regularization to ensure that the solution chosen by the optimization process is the best among the redundant solutions and specifically minimizes the time action $\tau$ (see Fig. \ref{fig:ik_model_and_reg}a). Scalar $\lambda$ is the regularization weight. The computation of $\mathcal{L}_{reg}$ will be discussed below.

The IK problem in the context of MASR does not only intend to find a configuration $\ve{q}$ corresponding to $\ve{x}_d$ but also minimize the joint and MA motion time from a current configuration $\ve{q}_{c}$. Hence, the model, illustrated in Fig. \ref{fig:ik_model_and_reg}b, takes in two inputs: the desired pose $\ve{x}_d$ and a current configuration $\ve{q}_{c}$. 
Regularization $\mathcal{L}_{reg}$ in \eqref{eq:loss_mse_reg} aims to minimize the motion from $\ve{q}_{c}$ to a configuration $\tve{q}_d=g_\psi(\ve{x}_d,\ve{q}_{c})$ that satisfies pose $\ve{x}_d$. The baseline regularization term $\mathcal{L}_{reg}$ often used in typical serial manipulators minimizes joint angle actions in the form
\begin{equation}
    \mathcal{L}_{reg}(\ve{q}_c,\tve{q}_d)=\sum_{i=1}^n |\theta_{c,i}-\tilde{\theta}_{d,i}| .
\label{eq:reg_term_theta}
\end{equation}
However, due to the unique design of the MASR, we construct a regularization term that minimizes action time. In other words, a modified regularization minimizes the joint rotation and MA moving time from $\ve{q}_{c}$ to the IK solution  $\tve{q}_d$, and is given by
\begin{equation}
    \mathcal{L}_{reg}(\ve{q}_c,\tve{q}_d)=\sum_{i=1}^n \left({\dot{\theta}}^{-1}|\theta_{c,i}-\tilde{\theta}_{d,i}|\cdot{\dot{d}}^{-1}|d_{c}-r_i|\right).
\label{eq:reg_term_q0}
\end{equation}
In such a form, the regularization term penalizes joints that take more time to activate due to long distance for the MA to reach and large rotation angles.
\begin{figure}[h]
\centering
\includegraphics[width=88mm]{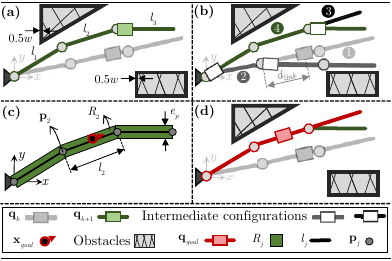}
\vspace{-0.6cm}
\caption{(a) Example of an environment containing two polygonal obstacles with typical configurations $\ve{q}_{k}, \ve{q}_{k+1}$. Fictitious margins, each of width $0.5w$, are added around obstacles. Here, $w$ is the actual links' width. Therefore, treating the arm as a series of lines and using a simple collision checker function is sufficient. (b) An illustration of the process of exerting an action $\ve{a}_k$ by the motion convention. Each step is numbered. The length to be moved will be $D_\ve{a}=d_{link}+2l_1+l_2$. (c) \texttt{OnGoal} function illustration. Given a goal $\ve{x}_{goal}$, the second rectangle, $R_2$, satisfies $R_2\cap \ve{x}_{goal}\neq\emptyset$. (d) Then, function \texttt{GoalFix} updates the new configuration $\ve{q}_{k+1}=\ve{q}_{goal}$ (red) and discards the old configuration $\ve{q}_{k+1}$ (green).}
\vspace{-0.3cm}
\label{fig:alg_illustration}
\end{figure}

Consider an example of a 5R MASR with link lengths of $l_{1,2,3}=l$ and $l_{4,5}=0.5l$ seen in Fig. \ref{fig:ik_model_and_reg}c. A motion from $\ve{q}_{c}=(\mathbf{0}^T, 1.5l)^T$ to another configuration $\tve{q}=(\alpha, 2\alpha, -3\alpha, -3\alpha, 2\alpha, 2.5l)^T$ that satisfies some $\ve{x}_d=f(\tve{q})$ is computed. Regularization term \eqref{eq:reg_term_q0} is $\mathcal{L}_{reg}(\ve{q}_c,\tve{q})=13.5\tau_{l}\tau_{\theta}$ 
where $\tau_{l} = l/\Dot{d}$ and $\tau_{\theta}=\alpha/\Dot{\theta}$ for some angle value $\alpha\in\mathbb{R}$. A straightforward approach to minimize these terms is to set $\Tilde{\theta}_{4}=\Tilde{\theta}_{5}=0$ since they are located after the position of MA and will not affect its pose. However, due to the high redundancy of the MASR, as demonstrated in Fig. \ref{fig:ik_model_and_reg}a, the configuration $\tve{q}$ can acquire alternative angles for $(\Tilde{\theta}_{1}, \Tilde{\theta}_{2}, \Tilde{\theta}_{3})$ that still satisfy $\ve{x}_d=f(\tve{q})$ potentially leading to additional minimization of $\mathcal{L}_{reg}$.

\vspace{-0.5cm}
\subsection{Data Generation} 
\label{sec:data_gen}

In order to train the IK-NN model $g_\psi$, a dataset $\mathcal{D}=\{(\ve{x}_{d,i})\}_{i=1}^N$ of $N$ MA pose samples is required. However, creating a balanced dataset with constant density that represents $\mathcal{W}$ is essential for predicting reachable MA poses and acquiring generalization. To achieve this, we propose the following steps. Due to the symmetry of MASR, the upper half workspace is defined as $\mathcal{W^{+}}=\{\mathbf{x} \in \mathcal{W} : \ y \geq 0\} \subset \mathcal{W}$ (see coordinate frame in Fig. \ref{fig:masr_and_scheme}). Subspace $\mathcal{W^{+}}$ is then divided into an occupancy grid denoted by $\ve{B}$. All cells in $\ve{B}$ are initially marked with zero values. A repetitive process samples random configurations $\ve{q}_i$ with computation of their corresponding poses $\ve{x}_i=f(\ve{q}_i) \in \mathcal{W^{+}}$. If a sample $\ve{x}_i$ falls into a position corresponding to cell $B_{a, b}$, the cell is set to one and the sample is added to $\mathcal{D}$. Hence, a cell $B_{a, b} = 1$ in $\ve{B}$ indicates the existence of a pose sample in the corresponding location. A new sample is discarded if it falls into an occupied cell marked with $B_{a, b} = 1$. Note that $\ve{x}_i\in\ve{B}$ indicates the existence of the unique MA pose in the cell while there can be a multitude of corresponding IK solutions with different arm configurations. These configurations are irrelevant to the training as we require only possible poses.

The above process of sampling and marking occupancy is repeated until density is sufficient. Grid density is heuristically considered sufficient if the size of vacancies fully surrounded by occupied cells are not larger than $\rho$ cells. Once density is sufficient, all samples in $\mathcal{D}$ are duplicated in order to obtain $\mathcal{W^{-}}$. Each point $\ve{x}_i\in\mathcal{D}$ is mapped to $\mathcal{W^{-}}$ by setting $y^{+}=-y^{-}$ and $\phi^{+}=-\phi^{-}$. All points are united into dataset $\mathcal{D}$ with $N$ pose samples. This data generation process ensures a uniform distribution of poses in $\mathcal{W}$ and sufficient coverage. Furthermore, during training and in each epoch, an initial configuration $\ve{q}_{c}$ is randomly generated for each $\ve{x}_{d,i}\in\mathcal{D}$. Hence, the model is introduced with poses and constantly changing initial configurations. Consequently, the IK-NN model learns to generalize to various initial configurations.

\section{Motion Planning with Obstacle-Avoidance}
\label{sec:mp}
\newcommand{\seqtime}{SeqsTimes}
\newcommand{\near}{Near}
\newcommand{\nearest}{Nearest}
\newcommand{\steer}{Steer}
\newcommand{\samplefree}{SampleFree}
\newcommand{\collfree}{CollisionFree}
\newcommand{\rewire}{RewireTree}
\newcommand{\ongoal}{OnGoal}
\newcommand{\goalfix}{GoalFix}
\newcommand{\steerik}{SteerIK}
\newcommand{\calcconfig}{CalcConfig}
\newcommand{\obsfree}{ObsFree}
\newcommand{\nearjoint}{NearJoint}
\newcommand{\concat}{Concat}
\newcommand{\find}{Connect}
\newcommand{\wirevg}{ImpWire}
\newcommand{\rand}{Rand}

\SetKwFunction{Fcalc}{\calcconfig}
\SetKwFunction{Fextract}{\seqtime}
\SetKwFunction{Fnear}{\near}
\SetKwFunction{Fnearest}{\nearest}
\SetKwFunction{Fsteer}{\steer}
\SetKwFunction{Fsample}{\samplefree}
\SetKwFunction{Fcoll}{\collfree}
\SetKwFunction{Fwire}{Wire}
\SetKwFunction{Frewire}{\rewire}
\SetKwFunction{Fcheck}{\ongoal}
\SetKwFunction{Fcopy}{Copy}
\SetKwFunction{Fgoal}{\goalfix}
\SetKwFunction{Fsteerik}{\steerik}
\SetKwFunction{Fobsfree}{\obsfree}
\SetKwFunction{Fnearjoint}{\nearjoint}
\SetKwFunction{Fconcat}{\concat}
\SetKwFunction{Ffindv}{\find}
\SetKwFunction{Fwirevg}{\wirevg}
\SetKwFunction{Frand}{\rand}

\SetKwIF{If}{ElseIf}{Else}{if}{}{else if}{else}{end if}%

\SetKwInput{KwInput}{Input}                
\SetKwInput{KwResult}{Output}              
\SetKw{KwTo}{in}

This section introduces the \mr algorithm. \mr is a modified version of RRT* adapted to motion planning for MASR. It is a hybrid obstacle avoidance motion planning algorithm that combines RRT* with the IK-NN model described in Section \ref{sec:ik}. \mr is designed to efficiently guide the RRT* search tree toward the goal pose $\ve{x}_{goal}$ while optimizing a path in the unique configuration space of MASR with faster convergence. 

\subsection{Node Formulation}
\label{subsub:base_rrt}
The baseline RRT* algorithm maintains a search tree $\mathcal{T}$ with potential trajectories. Each node $v\in\mathcal{T}$ has a parent node $u\in\mathcal{T}$ denoted by $v.parent=u$. The node $v$ consists of a configuration $v.\ve{q}\in\mathcal{C}_{free}$ and action $v.\ve{a}$ to move from $v.parent$ to $v$. Node $v$ also includes the total motion time $v.\tau$ taken to reach from the root node $v_0$ to $v$ calculated according to \eqref{eq:time_action}. In addition to tree $\mathcal{T}$, a subset $\mathcal{G}\subset\mathcal{T}$ is maintained during the search where nodes that reach $\ve{x}_{goal}$ by satisfying \eqref{eq:goal_set} are stored. 





\subsection{\mr}
\label{sub:base_rrt}

Algorithms \ref{alg:main_base}-\ref{alg:wire} depict the planning process of \mrr. In general, a search tree $\mathcal{T}$ is initialized with a start node $v_0$ comprising the initial configuration $\ve{q}_{init}$. For $N_c$ iterations, attempts are made to explore the configuration space and expand the tree towards the goal. Function \FuncSty{\samplefree} generates a random configuration $\ve{q}_{rand}\in \mathcal{C}_{free}$. Joint angles and MA position are sampled uniformly. Note that the RRT* includes a common goal biasing step where $\ve{q}_{goal}$ is sampled with some low probability instead of $\ve{q}_{rand}$. In function \FuncSty{\nearest}, the nearest node $v_{nearest}\in\mathcal{T}$ to $\ve{q}_{rand}$ is chosen for propagation having the minimal cost metric \eqref{eq:one_action}.  Propagation from $v_{nearest}$ is commonly performed in traditional RRT algorithms by function \FuncSty{\steer}. The function computes an action $\ve{a}$ to move from configuration $v_{nearest}.\ve{q}$ towards $\ve{q}_{rand}$ with a predefined maximum step size $\delta$. Hence, an action would be $\ve{a}_{new}=\delta(\ve{q}_{rand}-v_{nearest}.\ve{q})$. The function returns a new node $v_{new}$ with $v_{new}.\ve{q}=v_{nearest}.\ve{q}+\ve{a}_{new}$, action $v_{new}.\ve{a}=\ve{a}_{new}$ and the cost $v_{new}.\tau=v_{nearest}.\tau+c(\ve{a}_{new})$. 

Function \FuncSty{\steer} enables the expansion of the search tree toward random configurations in an attempt to finally reach the goal. In order to reach possible configurations that correspond to the goal pose $\ve{x}_{goal}$, we add a greedy step using the trained IK-NN model. With probability $p_c$, steering is done from $v_{nearest}$ towards the $\ve{x}_{goal}$ using the IK-NN model (Alg. \ref{alg:main_base}, line \ref{ln:iknn}). This step replaces the traditional goal biasing of RRT*. As described in Section \ref{sec:optimal_tran_method}, the solution of $g_\psi$ would provide a minimal motion configuration from $v_{nearest}.\ve{q}$ to a configuration $v_{new}.\ve{q}$ that is on the goal. However, any steering may be infeasible due to collision with obstacles. In order to prevent redundant computations, a subset $\mathcal{N}$ stores nodes where propagation from them to the goal was already checked using the IK-NN. Steering to the goal with IK-NN may add computational complexity while using a GPU can provide runtime compensation. It is important to note that, even if $p_c=1$, steering will be performed. Any $v_{nearest}$ that was already checked to reach the goal in a previous iteration is put in $\mathcal{N}$. Hence, it will not be checked again to reach the goal due to the second condition in Line \ref{ln:p_c} and will be propagated using \FuncSty{\steer} (Line \ref{ln:steer}). Moreover, the IK-NN model may output slightly erroneous approximations leading to propagation towards the goal but not reaching it. This expands the tree towards the vicinity of the goal and may also occur even with an ill-trained IK-NN model. 


Function \FuncSty{\collfree} checks whether motion between two configurations $\ve{q}_{nearest}$ and $\ve{q}_{new}$, based on the convention in Section \ref{sec:problem_def}, results in collision with obstacles. A collision checker for whether a configuration $\ve{q}$ is in $\mathcal{C}_f$ is assumed to exist. First, the algorithm checks if $\ve{q}_{nearest},\ve{q}_{new}\in\mathcal{C}_{free}$. If so, the algorithm rotates each joint according to the convention and validates that the intermediate configuration is in $\mathcal{C}_{free}$. If all configurations are collision-free, the function returns true. If motion from $v_{nearest}$ to $v_{new}$ is feasible, function \Ffindv in Algorithm \ref{alg:wire} attempts to improve connection of $v_{new}$ to the tree through a set of $N_n$ nearest neighbors $\mathcal{U}\subset\mathcal{T}$. The node in $\mathcal{U}$ that provides the minimum-cost path to $v_{new}$ based on \eqref{eq:time_action} is set to be the parent of $v_{new}$. Then $v_{new}$ is added to $\mathcal{T}$.

\begin{figure}[!t]
\removelatexerror
\begin{algorithm}[H]
\small
\caption{\mr - Main}\label{alg:main_base}
\SetKwInput{KwData}{Input}
\KwData{$\ \mathbf{q}_{init}, \ \mathbf{x}_{goal}, \ N_c, \ N_n.$} 
\SetKwInput{KwResult}{Output}
\KwResult{$\sigma^*$}
\SetKwInput{KwResult}{Initialization}
\KwResult{$v_0.\mathbf{q} \gets \mathbf{q}_{init}$, $v_0.\tau=0$, $\mathcal{T}\leftarrow {v_{0}}$, $\mathcal{G} = \emptyset$, $\mathcal{N}=\emptyset$}

\For{$N_c$ iterations}{
    $\mathbf{q}_{rand} \leftarrow$ \Fsample{} \\
    $v_{nearest}$ $\leftarrow$ \Fnearest{$\mathcal{T}, \ \mathbf{q}_{rand}$} \\ 
    $p$ = \Frand{$0,1$} \\
    
    \eIf{$p < p_{c} \ \wedge \ v_{nearest}\notin\mathcal{G} \cup \mathcal{N}$ \label{ln:p_c}}{
            $\mathcal{N} \leftarrow \mathcal{N} \ \cup \ \{v_{nearest}\}$ \\
            $v_{new}.\ve{q} \gets g_\psi(\ve{x}_{goal},v_{nearest}.\ve{q})$~~ \label{ln:iknn} \tcp{IK-NN} 
            $v_{new}.\tau \gets v_{nearest}.\tau+c(v_{new}.\ve{q}- v_{nearest}.\ve{q})$\\
        }
        {
        $v_{new}$ $\leftarrow$ \Fsteer{$v_{nearest}, \ \ve{q}_{rand}$} \label{ln:steer} \\
        }
    \If{\Fcoll{$v_{nearest}.\mathbf{q}$, \ $\mathbf{q}_{new}$}}{
        $\mathcal{U}$ $\leftarrow$ \Fnear{$\mathcal{T}, \ v_{new}.\ve{q}, \ N_n$} $\setminus \  \mathcal{G}$\\
        
        $v_{new}$ $\leftarrow$ \Ffindv{$\mathcal{U}, \ v_{nearest}, \ v_{new}$} \\
        
        $\mathcal{T}\gets\mathcal{T}\cup\{v_{new}\}$\\
        \eIf{\Fcheck{$\ve{x}_{goal}, \ v_{new}.\ve{q}$}\label{ln:goal_check}}{
            $v_{new} \leftarrow$ \Fgoal{$\ve{x}_{goal}, \ v_{new}$} \\
            $\mathcal{G}$ $\leftarrow$ $\mathcal{G} \  \cup \ \{v_{new}\}$
            } 
            { 
            \Frewire{$\mathcal{U} \setminus \{v_{new}.\mathrm{parent}\} , \ v_{new}$}\label{ln:rewire}}
        }
}
\Return{$\sigma^*\gets$\texttt{Best\_path}$(\mathcal{G})$}\\
\end{algorithm}
\vspace{-0.5cm}
\end{figure}



        
         
            
            
     
    

\begin{figure}[!t]
\removelatexerror
\begin{algorithm}[H]
\small
\caption{\FuncSty{\find}($\mathcal{U}, \ v_{near}, \ v_{new}$)} 
\label{alg:wire}

$v_{min} \gets v_{near}$; \ $\tau_{min} \gets v_{new}.\tau$; \\

\For{$u$ \KwTo $\mathcal{U}$}{
        
         $\ve{a}_{new}\gets v_{new}.\ve{q} - u.\ve{q}$\\
         $\tau \leftarrow u.\tau + c(\ve{a}_{new})$\\
         
        \If{$\tau<\tau_{min} \ \wedge$ \Fcoll{$u.\ve{q}, v_{new}.\ve{q}$}}{
        
                    $v_{min} \gets u;\
                    \tau_{min} \gets \tau$\\
                    }

    }
    $v_{new}.parent \gets v_{min}$; \ $v_{new}.\tau \gets \tau_{min}$\\  
    \Return{$v_{new}$}
\end{algorithm}
\vspace{-0.5cm}
\end{figure}

Once $v_{new}$ was added to the tree, the algorithm checks whether $f(v_{new}.\ve{q})\in\mathcal{W}_{goal}$ 
in Function \FuncSty{\ongoal} (Line \ref{ln:goal_check}). However, it may be possible that some point on the arm is in $\mathcal{W}_{goal}$ while the MA with the gripper is not. Hence, function \FuncSty{\ongoal} models the links of the serial arm as elongated rectangles $\{R_1, \dots, R_n\}$, where each rectangle $R_j$ corresponds to link $l_j$ and has a width equal to $e_{p}$. The function returns true if some rectangle $i$ satisfies $R_i\cap\mathcal{W}_{goal}\neq\emptyset$ (see Fig. \ref{fig:alg_illustration}c). If true, function \FuncSty{\Fgoal} does two corrections to $v_{new}$. First, it updates the position $d_{new}$ in $v_{new}.\ve{q}$ of the MA to be in $\mathcal{W}_{goal}$ without modifying the joint angles. Second, the function checks whether there are redundant joint movements in the propagation from $v_{new}.parent$ to $v_{new}$. Joints $j=i,\ldots,n$ for some $i>1$ such that $r_j>d_{goal}$ have no reason to be rotated when transitioning between the nodes since the arm and MA will be on the goal either way. Hence, joint angles $(\theta_i,\ldots,\theta_n)_{new}\in v_{new}.\ve{q}$ would be equal to $(\theta_i,\ldots,\theta_n)_{parent}\in v_{new}.parent.\ve{q}$  (see Fig. \ref{fig:alg_illustration}d). In addition, the action time value $v_{new}.\tau$ would be reduced as fewer joints are required to be actuated. After fixing the new node, it is added to the goal set $\mathcal{G}$ as a possible solution.

If $v_{new}$ did not reach the goal, a rewiring step is performed in order to maintain optimal paths in the tree. With $\mathcal{U} \subset \mathcal{T}$, function \FuncSty{\rewire} in Line \ref{ln:rewire} iterates through the vertices $u \in \mathcal{U}$. If a connection of $v_{new}$ with some $u$ is collision-free and $v_{new}.\tau+c(u.\ve{q}-v_{new}.\ve{q})<u.\tau$, the function updates the parent such that $u.parent=v_{new}$. This step effectively rewires the tree structure and guarantees convergence to optimum. Finally, function \texttt{Best\_path} selects $v_{end}\in\mathcal{G}$ which is the solution of $\min_{v.\tau}\mathcal{G}$ and returns the optimal path $\sigma^*$ from $v_0$ to $v_{end}$.

\vspace{-0.3cm}
\section{Experiments}
\label{sec:exp}
\newcommand{\RNum}[1]{\uppercase\expandafter{\romannumeral #1\relax}}

\begin{table*}[]
\tabcolsep=0.176cm
\caption{Comparative results for different IK models}
\vspace{-0.2cm}
\label{tab:hp}
\centering
\begin{tabular}{ccccccccccc}
\hline
 \multirow{2}{*}{Model} &  \multirow{2}{*}{Reg.} & \multirow{2}{*}{$\lambda$} & Size of & Learning & Batch & \multicolumn{2}{c}{Mean errors} & Success & Action & Runtime \\\cline{7-8}
   & & &  hidden layers & rate  & size &  $\Delta{p}$ {(}mm{)} & $\Delta{\phi}$ {(}deg{)} & rate (\%) & time (s) & (s) \\ \hline
   
 I & \eqref{eq:reg_term_theta}  & 0.002 & [70,50,30,20] & $10^{-4}$ & 500 & 3.63 & 3.09 & 86 & 18.77 &  $\num{4e-4}\pm\num{4e-4}$  \\ 
  II &\eqref{eq:reg_term_q0} & 0.001 & [120,100,50,30] & $10^{-4}$ & 500 & 3.52 & 3.43 & 87 & 15.24 & $\num{5e-4}\pm\num{4e-4}$ \\ \hline
  IK w/ NR & \multicolumn{4}{c}{$m=10^3$} &  & 2.31  & 0.42  & 97 & 16.59 & 0.24 $\pm$ 0.007\\ 
  IK w/ NR & \multicolumn{4}{c}{$m=10^4$} &  & 2.63  & 1.21 & 99 & 16.03  & 2.37 $\pm$ 0.065 \\ \hline 
\end{tabular}
\vspace{-0.5cm}
\end{table*}
 
This section presents experiments of the methodologies and algorithms, and practical implementations using the MASR experimental setup. 


\subsection{Experimental Setup}
\label{subsec:experimental_setup}
Towards experiments on a real MASR, we evaluated \mr on an $n=5$ arm with $l=0.8~m$. Link lengths are $l_1=l_2=l_3=0.2~m$ and $l_4=l_5=0.1~m$. The shorter links enable improved dexterity when reaching targets with the distal extremity and demonstrate the effectiveness of MASR-RRT* with an arbitrary, non-uniform arm configuration. All joints $j=1,\ldots,n$ can move within the range of $[-\theta_{j}^b, \theta_{j}^b]=[-50^\circ, 50^\circ]$. Also, the goal region defined in \eqref{eq:goal_set} has bounds $e_{p}=8~mm$ and $e_{\phi}=4^{\circ}$. The velocities of the MA and joints are controlled using PID controllers and maintained at $\dot{d}=100~mm/s$ and $\dot{\theta}=0.28~rad/s$, respectively.


\subsection{IK-NN Evaluation}
\label{subsec:ik_eval}

IK-NN is evaluated where the MA must reach a desired pose $\ve{x}_d\in SE(2)$ while comparing between the performances of regularization terms \eqref{eq:reg_term_theta} and \eqref{eq:reg_term_q0}. Dataset $\mathcal{D}$ was collected as discussed in Section \ref{sec:data_gen} by generating a grid of size $1,800\times1,600$ over $\mathcal{W}^+$. With a density constraint of $\rho=10$ cells and approximately $\num{1.16e6}$ random samples, dataset $\mathcal{N}$ reached $N=\num{0.5e6}$ samples uniformly distributed in $\mathcal{W}$. We reduce the input of the goal pose to $x_{goal}$, $y_{goal}$ and $\phi_{goal}$ while the trained models output the estimated configuration of dimension $n+1=6$. The hyper-parameters of each model were optimized to provide minimal loss \eqref{eq:loss_mse_reg}. The optimal hyper-parameters are summarized in Table \ref{tab:hp} and include regularization weight $\lambda$, learning rate, batch size and network architecture. All models yielded networks with four hidden layers. The training of the models was done over 1,000 epochs. We also compare IK-NN to numerical IK solution. For each IK query, $m$ solutions are acquired using a Newton-Raphson (NR) solver from the Robotics-Toolbox \cite{haviland2023dkt1}. The $m$ solutions are divided into $n$ groups where each group solves for an arm with the first $n-i$ joints ($i=1,\ldots,n$) and an additional prismatic joint simulating the MA at the last link. 
The solution with the minimum MA action time \eqref{eq:reg_term_q0} is picked out of the $m$ solutions.

All IK models are evaluated on 5,000 test trials moving from random initial configurations $\ve{q}_{0}$ to desired random MA poses $\ve{x}_d$. The random configurations and poses were sampled from a uniform distribution. In addition to the mean position and orientation errors over the training and test data, Table \ref{tab:hp} reports the success rate of planning to reach $\ve{x}_d$ with the accuracy defined by \eqref{eq:goal_set}. IK with NR is evaluated using $m=10^3$ and $m=10^4$. For IK-NN, the results show sufficient mean accuracy and high success rates in reaching the goals within the defined error bounds. 
While the accuracies of regularizations \eqref{eq:reg_term_theta} and \eqref{eq:reg_term_q0} are roughly similar, the former does not have the ability to minimize action time. Hence, the mean action time for \eqref{eq:reg_term_q0} is lower. 
For numeric IK, the pose errors are lower with higher success rates, while not being able to sufficiently minimize the action time. In addition, the runtime for an IK query is at least two orders of magnitude higher than IK-NN. This high runtime for a single IK query yields an extremely long planning time and does not enable real-time planning. We note that the accuracy of IK-NN can be improved at the cost of increasing the number of hidden layers and runtime.

\subsection{\mr Evaluation}

With the trained IK-NN model, we now evaluate the \mr algorithm compared to the baseline. The baseline RRT* is the one where there is no use of the IK-NN and propagation is done conventionally with the \texttt{Steer} function. Also, goal biasing for RRT* is set with probability 0.1. First, the planning algorithm was evaluated for cost convergence and ability to find solutions with regard to the number of iterations and probability $p_c$. For analysis, 300 planning environments were generated where in each, the initial configuration $\ve{q}_{init}$ and goal pose $\ve{x}_{goal}\in SE(2)$ were uniformly randomized. 
Also, up to four polygonal obstacles were randomly generated such that they cover up to 30\% of the workspace. All 300 environments are known to have a path solution. 
\begin{figure}[]
\centering
\includegraphics[width=\linewidth]{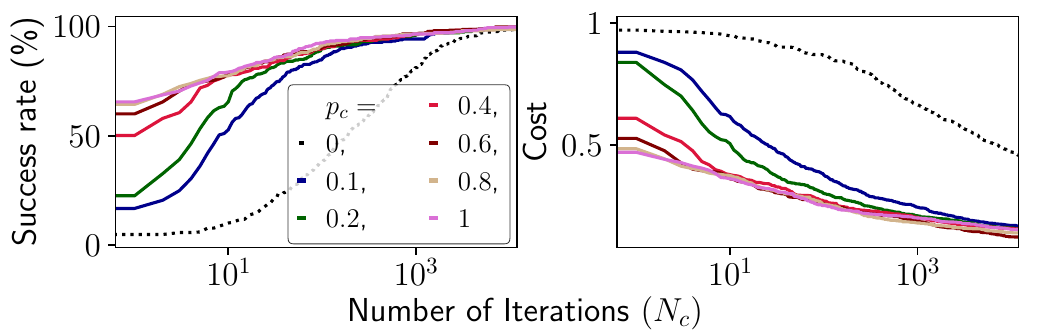}
\vspace{-0.8cm}
\caption{Success rate and cost value for 300 random environments, with regard to the number of planning iterations $N_c$ and probability $p_c$.}

\label{fig:obs_cost_sr}
\vspace{-0.5cm}
\end{figure}
\begin{figure}[]
\centering
\includegraphics[width=\linewidth]{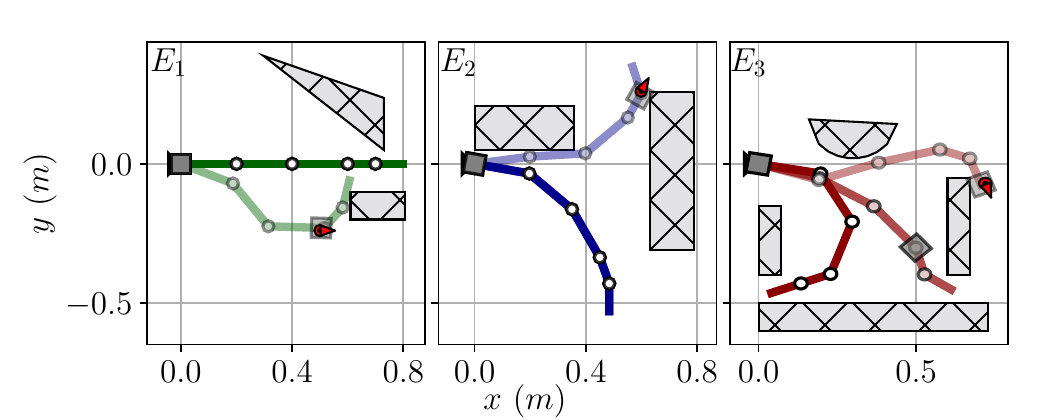}
\vspace{-0.9cm}
\caption{Three environments, $E_1$, $E_2$ and $E_3$, used for analysis where each having polygonal obstacles. The non-faded configurations are the initial ones. Similarly, the faded configurations illustrate the optimal ones reaching the corresponding goals after planning with MASR-RRT*.}
\label{fig:env}
\vspace{-0.6cm}
\end{figure}

Fig. \ref{fig:obs_cost_sr} presents the success rate and cost with regard to the number of iterations $N_c$ and probability $p_c$ of using the IK-NN. Note that the algorithm with $p_c=0$ is the baseline RRT* with regular goal biasing. 
In general, the results show a success rate and cost improvement with any choice of $p_c>0$. Without using the IK-NN, the convergence is much slower and requires many iterations. The mean planning time for RRT* over 1,000 iterations is 14.2 seconds. For MASR-RRT* with $p_c=0.2$, $p_c=0.6$ and $p_c=1$, the planning time over the same number of iterations is 12.9, 15.1 and 15.9 seconds, respectively. Since the addition of IK-NN with an increased probability $p_c$ provides a high success rate and low cost at the early stages of planning, the planning time is significantly reduced. Hence, the use of \mr over the baseline is justified and advantageous. 

\begin{figure}[]
\centering
\includegraphics[width=\linewidth]{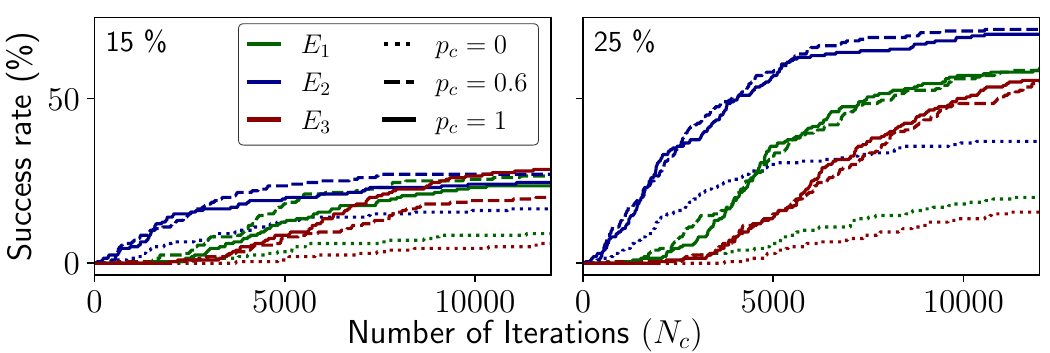}
\vspace{-0.5cm}
\caption{Success rate for reaching below (left) 15th and (right) 25th percentile of the cost range for $E_1$-$E_3$, with regards to number of iterations.}
\label{fig:env_plots}
\vspace{-0.5cm}
\end{figure}

In the next experiment, the performance of \mr is analysed on three environments, $E_1$, $E_2$ and $E_3$, seen in Fig. \ref{fig:env}. The latter environment is more challenging than the former two as the arm must first navigate through the bottom-right opening prior to reaching the designated goal. An experiment for each environment consists of 200 planning trials executed from the same initial configuration to the goal $\ve{x}_{goal}$. Each trial enabled up to $N_c=12,000$ iterations and had $N_n=7$ neighbors in $\mathcal{U}$. Also, a constant seed was used in both \mr and baseline algorithms for a fair comparison. During each trial and once some solution was found, the costs along the iterations are recorded. Then, we obtain the maximum and minimum cost for the 200 trials and normalize the cost to be in the range of $[0,1]$. Fig. \ref{fig:env_plots} shows the success rate for reaching below 0.15 and 0.25 of the normalized cost range (i.e., below 15th and 25th percentile, respectively), with regards to the number of iterations. 

The results show that \mr is at least equivalent to the baseline RRT* (when $p_c=0$) and, in most cases, superior. The results emphasize the dominance and performance benefits of \mrr. A key advantage of MASR-RRT* is in the IK-NN based goal biasing compared to the traditional goal biasing of RRT*. While RRT* searches for specific joint configurations, MASR-RRT* with IK-NN is able to explore a multitude of joint configurations for a desired MA pose. In addition, IK-NN naturally offers motion perturbations for expanding the search tree that reduce the action time. Hence, the planning is expedited.

\begin{table}[h]
\vspace{-0.5cm}
\caption{Simulation and Demonstration Results} 
\label{tb:sim_demo_results}
\centering
\begin{tabular}{ccccc}
\hline
\multirow{2}{*}{Env.} & Scenario & \multicolumn{2}{c}{Pose errors} & Cost \\\cline{3-4}
                     & & $\Delta{p}$ (mm) & $\Delta{\phi}$ ($^\circ$) & $\tau \ (s)$               \\ \hline
\multirow{2}{*}{$E_1$} & Planned & 7.1 & 1.9 & 46.16  \\
                       & Real robot & 11$\pm$3.1 & 1.2$\pm$ 0.8 & 48.78$\pm$1.81 \\ \hline
\multirow{2}{*}{$E_2$} & Planned & 4.75 & 0.7 & 39.46 \\
                       & Real robot & 10.85$\pm$5.32 & 3.4$\pm$ 3.8& 42.32$\pm$2.23 \\ \hline
\end{tabular}
\vspace{-0.8cm}
\end{table}

\subsection{MASR Demonstration}
\label{subsec:rrt_imp}
The optimal solutions of \mr for environments $E_1$ and $E_2$ within $SE(2)$ are next evaluated with a real MASR robot. The paths are tested on the MASR experimental setup (Fig. \ref{fig:masr}) with $n=5$ detailed in Section \ref{subsec:experimental_setup}. We evaluate the real-time performance while comparing planned paths to desired goal poses with real robot roll-outs. Real robot poses are measured using a motion capture system. Videos of the experiments can be seen in the attached video.

\begin{figure*}
\centering
\begin{tabular}{c}
\includegraphics[width=\linewidth]{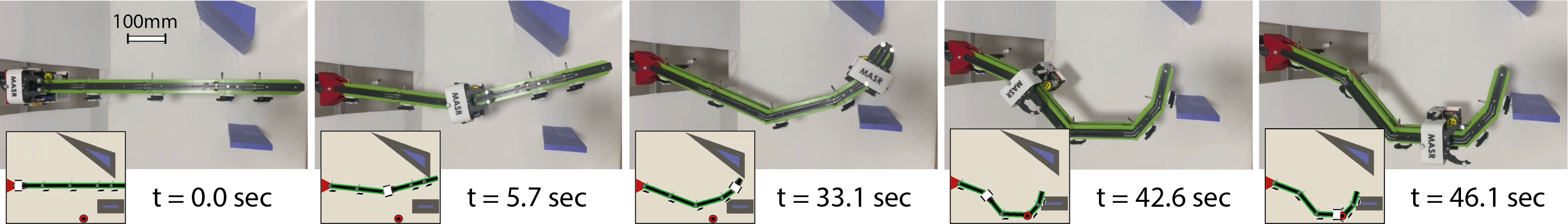}\\
\includegraphics[width=\linewidth]{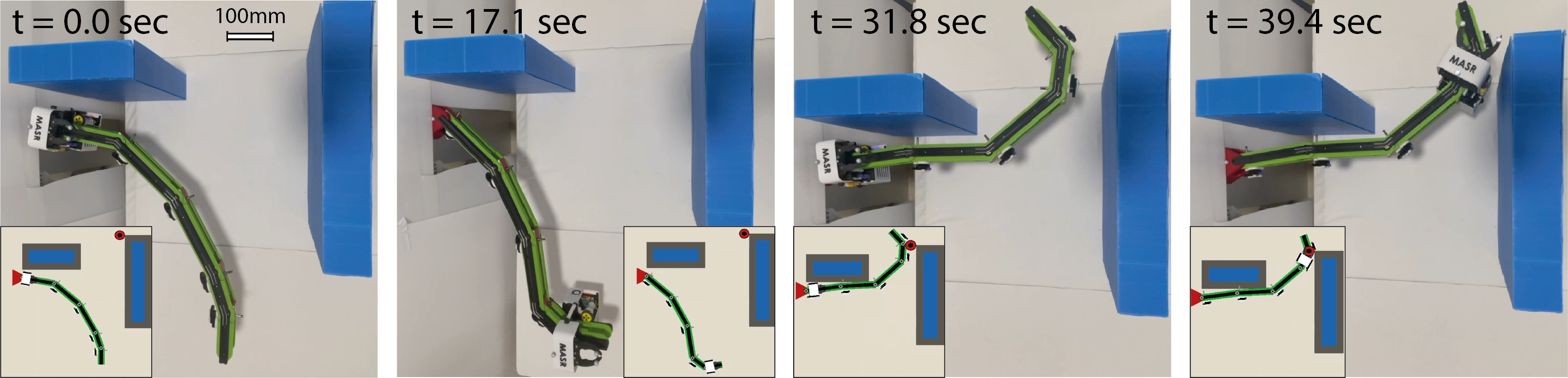}
\end{tabular}
\vspace{-0.3cm}
\caption{Snapshots of path roll-outs along with illustrations of the planned paths for environments (top row) $E_1$ and (bottom row) $E_2$.}
\label{fig:ex_env2}
\vspace{-0.4cm}
\end{figure*}

Table \ref{tb:sim_demo_results} presents the mean pose errors, according to \eqref{eq:goal_set}, of the MA over ten roll-out trials of the planned paths with the real robot. Pose error is decomposed to the position and orientation errors of the MA with respect to the goal. The yielded accuracy is sufficiently small while the marginal errors are due to fabrication uncertainties of the robot. The runtime costs are also reported in the table. The mean cost deviations from the planned paths in environments $E_1$ and $E_2$ are $2.62$ and $2.86$ seconds, respectively. These are mainly influenced by the dynamics of the PID controller which are not modeled in \eqref{eq:time_action}-\eqref{eq:one_action}. However, the results validate the robot's ability to follow planned paths and accurately reach the desired poses of the goal. Roll-out example trials with the real robot in environments $E_1$ and $E_2$ are seen in Fig. \ref{fig:ex_env2}
, along with illustrations of the planned motions. The roll-outs show successful tracking over planned paths. A demonstration of the MASR moving along a planned path to grasp and retrieve a balloon can be seen in Fig. \ref{fig:masr} and in the accompanying video. The success rate for reaching and grasping the balloon was 100\% out of five trials.

\section{Conclusions}
In this letter, we have addressed the motion planning problem for the MASR. The \mr algorithm was proposed and is a modified RRT* specifically designed for the unique kinematics of MASR. The main component is a data-based IK solver which provides a minimal traverse for the current configuration without considering obstacles. The IK-NN model exhibited high accuracy for reaching desired pose goals. Along with IK-NN, the \mr algorithm minimizes the action time of the robot while considering obstacle-free paths. \mr is shown, through a set of evaluations, to outperform the standard RRT* in finding better solutions and converging in less iterations. Experiments on a real robot in two environments validate the planning results and show successful reach to targets.

MASR is an advantageous mechanism in applications requiring high-dimensionality with low-cost and simplicity. Our work provides an algorithmic tool for planning efficient paths for MASR. Future work could consider the motion planning for a MASR that spans the 3D space. Furthermore, one could explore the integration of MASR with various sensors and data-driven decision-making algorithms. For instance, reinforcement learning can be used for MASR to autonomously learn to operate in an unknown environment with minimal sensing. The robot will constantly map its proximity and re-plan the shortest path to the goal with partial information.


\bibliographystyle{IEEEtran}
\bibliography{ref}

\end{document}